# De-identification of medical records using conditional random fields and long short-term memory networks

Zhipeng Jiang[*]      hit.jiang@hotmail.com
Chao Zhao[*]      zhaochaocs@gmail.com
Bin He      hebin_hit@hotmail.com
Yi Guan[†]      guanyi@hit.edu.cn
Jingchi Jiang      jiangjingchi0118@163.com
*School of Computer Science and Technology*
*Harbin Institute of Technology*
*Harbin, Heilongjiang, 150001, CHN*

## Abstract

The CEGS N-GRID 2016 Shared Task 1 in Clinical Natural Language Processing focuses on the de-identification of psychiatric evaluation records. This paper describes two participating systems of our team, based on conditional random fields (CRFs) and long short-term memory networks (LSTMs). A pre-processing module was introduced for sentence detection and tokenization before de-identification. For CRFs, manually extracted rich features were utilized to train the model. For LSTMs, a character-level bi-directional LSTM network was applied to represent tokens and classify tags for each token, following which a decoding layer was stacked to decode the most probable protected health information (PHI) terms. The LSTM-based system attained an i2b2 strict micro-$F_1$ measure of 89.86%, which was higher than that of the CRF-based system.

## 1. Introduction

The Electronic Health Record (EHR) is the systematized collection of electronically stored health information of patients in digital format [1]. It consists of a large amount of medical knowledge, and is a novel and rich resource for clinical research. A limitation of the large-scale use of EHR is the privacy of information contained in the text. To protect the privacy of patients and medical institutions, the US Congress passed the Health Insurance Portability and Accountability Act (HIPAA) in 1996. HIPAA defines 18 kinds of protected health information (PHI) that must be removed before the EHR can be reused, such as names, all geographic subdivisions smaller than a State, and so on[1]. The

---

[*]. These authors contributed equally to this work.
[†]. corresponding author
1. https://www.hipaa.com/hipaa-protected-health-information-what-does-phi-include/ lists all PHIs in the HIPAA

i2b2 (Informatics for Integrating Biology and the Bedside) Center defines more types of PHI based on the HIPAA-PHI categories. The removal of PHI information from clinical narratives is called *de-identification*. However, manual de-identification is time consuming, expensive, and ineffective. To explore the possibility of automatic de-identification approaches using natural language processing (NLP), i2b2 held its first clinical narrative de-identification event in 2006 [2], and again in 2014 (The 2014 i2b2/UTHealth NLP Shared Task 1) [3], and 2016 (CEGS N-GRID 2016 Shared Task 1 in Clinical NLP) [4]. Most participants of the events proposed solutions to this problem using machine learning algorithms, whereas rule-based methods were also presented.

According to the results, the highest-ranking team attained an i2b2 strict micro-$F_1$[2] of over 90%. Although certain PHI can not be de-identified by an automatic system, studies have shown that it is sufficient for preventing re-identification from these processed records [5, 6]. These studies together confirmed the efficiency of the automatic de-identification systems.

In this paper, we describe two de-identification systems utilized in CEGS N-GRID 2016 Shared Task 1 based on conditional random fields (CRFs) and long short-term memory networks (LSTMs). We also contrast the principle and performance of these two systems, and analyze the identification errors. The remainder of this paper is structured as follows: In Section 2, we give a brief introduction of recent models used for named entity recognition (NER) and medical narrative de-identification. Section 3 describes the general pipeline of this task, and Sections 4 and 5 provide details of the principles and implementation of CRFs and LSTMs, respectively. We then report the evaluation of our systems on the CEGS N-GRID 2016 Shared Task 1 dataset in Section 6, and provide results and discussion in Section 7 and 8, respectively. Our conclusions and directions for future studies are presented in Section 9.

## 2. Related Work

From the perspective of NLP, de-identification is an NER task. NER was first introduced at the Sixth Message Understanding Conference (MUC-6) [7], and has developed rapidly in the 20 years since. Many statistical learning algorithms have been applied to it, such as hidden Markov networks (HMMs) [8], CRFs [9], support vector machines (SVMs) [10]. These methods all depend on several features and regard the problem as a tagging process, which is the classification of each token over text sequences. The common features include word-level, list-level (i.e., dictionary features), and document-level features. The CRF is the most commonly used model in general NER tasks because of its theoretical advantage and experimental efficiency [11]. In the past two i2b2 de-identification tasks in 2006 and 2014, the best systems were based on CRFs. In the 2014 task, [12] identified the word-token, context, orthographical, sentence-level, and dictionary features to train a CRF model, and achieved the highest F-measure of 93.6% of all participants. Moreover, manually derived post-processing approaches are often used, and can yield considerable improvement in some PHI categories, like DATE and HOSPITAL.

---
2. The evaluation measures are introduced in Section 6.4

In spite of their good performance, a problem with these approaches is that performance is highly dependent on the extracted features. The quality of the features relies heavily on the experience of researchers and their familiarity with the data. In recent years, the rise of representation learning [13] methods has brought new vitality to NER tasks. Representation learning attempts to extract efficient features directly from data, and then applies deep neural networks [14], such as convolutional neural networks (CNNs) and recurrent neural networks (RNNs), to compose the features. Features are then transformed layer by layer using non-linear functions to fit the intricate structures of the data. Better performance is obtained than that by CRFs when the training data is abundant. Many modified and combined deep neural networks have been applied to tagging tasks, from the simplest feed-forward neural networks [15] to long short-term memory (LSTM) networks [16] and various combinations, such as LSTM-CRF [17], LSTM-CNNs [18], CNN-LSTM-CRF [19], and so on. [20] proposed a character-level bidirectional LSTM-CRF architecture and claimed to obtain state-of-the-art NER results in standard evaluation settings. [21] transferred this work to EHR de-identification tasks and obtained an $F_1$ measure of 97.85% on the i2b2 2014 dataset, higher than that of the best CRF-based approach in [12] (93.6%).

In addition to statistical approaches and deep learning methods, rule-based methods are helpful for NER, although they are usually adopted as a deliberately weakened component in many academic papers [22]. Such rules include regular expressions, domain dictionaries, and a series of hand-crafted grammatical, syntactic, and orthographic patterns.

## 3. De-identification Pipeline

Although many machine learning approaches are available for NER, they follow the general processing procedure of pre-processing, tagging, and post-processing, which is also used in our two de-identification systems. Pre-processing is indispensable when the data are not as clean as expected. After pre-processing, we use an algorithm to tag the entities through annotated data[3], or use several models to boost the results. During post-processing, hand-derived rules are applied to correct potential tagging errors and find more missing entities. It is not indispensable, but can help improve the accuracy of the system in many cases. Since we had limitations of time during the task, we did not introduce a post-processing module. This section gives an overview of pre-processing and tagging modules in our two systems.

---
3. In this paper, we focus on supervised NER approaches

### 3.1 Pre-processing

In pre-processing, we focus on tokenization and sentence detection. Tokenization is necessary because some separators between entity tokens and ordinary tokens may be missing. Without these segments, the tagger cannot accurately detect the boundaries of entities. For example, in the phrase "09/14/2067CPT Code," "09/14/2067" is the date entity, but "CPT Code" is not. Without tokenization, the tagger would either recognize "09/14/2067CPT" as an entity or not, and neither is correct. In this case, extent or missing errors[4] occur.

Some kinds of cases can be tokenized easily through regular expressions, such as the above example. In other cases, a token dictionary is needed. For simplicity, we only tokenize text by regular expressions listed in Table 1.

Table 1: Regular expressions used for tokenization.

| Regular expression | Original token | After tokenization | Comment |
| --- | --- | --- | --- |
| [A-Za-z][0-9] | a26 yo man | a␣26 yo man | digit |
| [0-9][A-Za-z][A-Za-z]+ | 10/6/2098SOS | 10/6/2098␣SOS | |
| [A-Z]{3,}[a-z]{2,}+ | USMeaningful | US␣Meaningful | uppercase |
| [a-z][A-Z] | WhalenChief | Whalen␣Chief | |
| \d{1,2}([/-])(\d{1,2}(\1))?\d{2,4} | 09/14/2067CPT | 09/14/2067␣CPT | DATE |
| \D\d{3}\D{0,2}\d{3}\D{0,2}\d{4} | 109 121 1400Prior | 109 121 1400␣Prior | PHONE |
| \w+@\w+\.[a-z]+ | hcuutaj@bdd.comOther | hcuutaj@bdd.com␣Other | EMAIL |

Sentence detection is another part of pre-processing. Records must be separated as sentences to feed into models because neither CRFs nor LSTMs can receive sentences that are very long. Detection only according to punctuation can cause problems. For example, if "Dr. Vincent" is separated according to the period, "Vincent" becomes the first word of the sentence and "Dr.," which is an important identifier of the DOCTOR category is lost. Sentence boundaries can be detected using either rule-based methods or machine learning-based methods[23]. In this study, we detected the boundaries using the OpenNLP sentence detector[5], which is a supervised toolkit. We only used the officially provided detection model.

---

4. We discuss the type of errors in Section 8.2
5. https://opennlp.apache.org/documentation/1.5.3/manual/opennlp.html#tools.sentdetect

## 3.2 Tagging

We used the BIOEU tagging schema for this task. It tagged the **B**eginning, **I**ntermediate, and **E**nd parts of the entities, as well as the **O**utside of a named entity. If an entity consisted of only one token, it was simply tagged as of **U**nit length.

Like other classification problems, tagging relies on feature extraction. The traditional features of text are indicator functions. These features are really flexible and have been shown to be efficient. However, they face two problems. First, these features are handcrafted, and feature templates need to be re-designed when handling new data. Second, these features are numerous and sparse. This large number of features leads to more parameters and higher computational cost. It also restricts the size of the context windows during feature extraction.

The emergence of representation learning and deep neural networks led to the introduction of new feature extraction approaches. Researchers tried to learn the distributed representation of each token, called word vectors, or word embedding [24]. This kind of representation maps discrete words as vectors in a continuous low-dimensional vector space. That is to say, for each word $w_i$, there is a map $\phi(\cdot)$ to ensure that $\phi(w_i) = \mathbf{w}_i \in \mathbb{R}^{d_e}$, where $d_e$ is the number of dimensions of the embedding. The word embeddings are trained according to the contexts of the words, and words with similar semantic meanings are mapped into nearby representations. In this way, language is converted into signals that can be numeralized and computed, just like sounds and images. The vectors of each word can be directly regarded as features, and $d_e$ is usually between 50 and 1,000, which is much smaller than in one-hot representation. The difference between word embeddings and other signals is that the operations for these vectors, such as translation, rotation, scaling and superposition, have no actual meanings, or their meanings are still unknown. However, to combine the representation of words into higher-level representations, like the representations of sentences, the word embeddings must be manipulated. Thus, different kinds of neural networks are used to combine word embeddings, including CNNs, RNNs, and recursive neural networks.

## 4. CRF-based Model

### 4.1 Principles

The CRF is a kind of undirected probabilistic graphical model (PGM) that consists of two groups of variables, $X$ and $Y$. If the value of $X$ is given, the probability of the value of $Y$ can be calculated. In NLP, we focus on linear-chain CRF, which was first introduced in [25]. The representation and inference part of linear-chain CRFs can be found in Appendix B.

The pipeline of our CRF-based system is shown in Figure 1. The CRF model was implemented using the CRF++ toolkit[6]. The details of feature generation is given in next section.

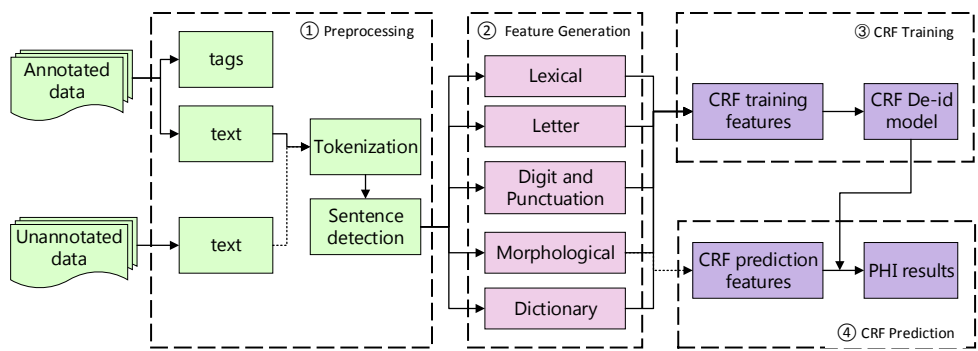

Figure 1: The pipeline of CRF-based tagging system. The raw text is parsed from training data, tokenized, and split into sentences. Five categories of features are generated for each token. These features are then fed into CRF++ toolkit to obtain the de-identification tagging model. The same pre-processing and feature generation steps are applied to test data, then the trained model would label each test token according to their features. The labels are decoded and the system would output the final PHI results.

### 4.2 Feature generation

A large number of rich features were extracted to feed the CRF classifier, including lexical, orthographic, morphological, and dictionary features. All features of the given token within its ±2 context window were considered. More details of the features are listed in Table 2. The POS and chunk features were obtained by utilizing OpenNLP POS tagger and chunker[7]. The dictionaries used in the system were collected from the training data as well as Wikipedia.

It is not the case that the more features, the better the performance. Some features may introduce noise and degrade performance. To avoid bad features, we selected them in a greedy way. That is, we added features in turn and evaluated the tagging results. Once performance degraded, we discarded the given added feature. This approach cannot guarantee optimal feature selection, but can reduce time complexity from $O(2^n)$ to $O(n)$ with ease. The contribution of each feature sub-category to the final performance is analyzed in Section 7.1.

---

6. https://taku910.github.io/crfpp/
7. https://opennlp.apache.org/

Table 2: The features utilized at the CRFs-based system.

| Category | Features | Feature Instantiations of "Vincent" |
|---|---|---|
| Lexical | lowercase | vincent |
|  | word lemma | vincent |
|  | POS tag of the token | NNP |
|  | Chunk tag | I-NP |
|  | Long Shape of the token | Aaaaaaa |
|  | Length of the token | 7 |
| Letter | Whether the token contains a letter | 1 |
|  | Whether the token contains a capital letter | 1 |
|  | Whether the token begins with a capital letter | 1 |
|  | Whether all characters in the token are capital letters | 0 |
| Digit and Punctuation | Whether the token contains a digit | 0 |
|  | Whether all characters in the token are digits | 0 |
|  | Whether the token contains a punctuation character | 0 |
|  | Whether the token consists of letters and digits | 0 |
|  | Whether the token consists of digits and punctuation characters | 0 |
| Morphological | First two characters of the token | Vi |
|  | Last two characters of the token | nt |
|  | First three characters of the token | Vin |
|  | Last three characters of the token | ent |
|  | First four characters of the token | Vinc |
|  | Last four characters of the token | cent |
| Dictionary | Whether the lowercase of the token is in the "profession" dictionary | 0 |
|  | Whether the lowercase of the token is in the "city" dictionary | 0 |
|  | Whether the lowercase of the token is in the "country" dictionary | 1 |
|  | Whether the lowercase of the token is in the "state" dictionary | 0 |

## 5. LSTM-based Model

LSTM is a special type of RNN. It utilizes word embeddings as inputs. The embedding of the given tokens are then combined with the context embeddings by the LSTM layer, which yields the new hidden representation of the tokens. Finally, the hidden representations are used directly for classification. These three steps are discussed in the following subsections. Figure 2 shows the architecture of the LSTM networks used in the task.

### 5.1 Long short-term memory networks

RNN is one way to combine a sequence of word embeddings $\mathbf{x}_{<1:t>}$ to an embedding $\mathbf{h}_t \in \mathbb{R}^{d_h}$. The combination is defined using the recurrent formula

$$\mathbf{h}_t = \tanh(\mathbf{W}\mathbf{x}_t + \mathbf{U}\mathbf{h}_{t-1} + \mathbf{b}), \tag{1}$$

where $\mathbf{x}_t \in \mathbb{R}^{d_e}$ is the word embedding of $t$-th word, $\mathbf{W} \in \mathbb{R}^{d_h \times d_e}$, $\mathbf{U} \in \mathbb{R}^{d_h \times d_h}$, $\mathbf{b} \in \mathbb{R}^{d_h}$ are the weight and bias parameters that to be learned, and the initial condition is $\mathbf{h}_0 = \mathbf{0}$.

After feeding all $\mathbf{x}_{<1:t>}$ into the above formula, we obtain $\mathbf{h}_t$, which contains not only the given token, but also the previous context as well, and can be used as a new representation of $t$-th word $w_t$.

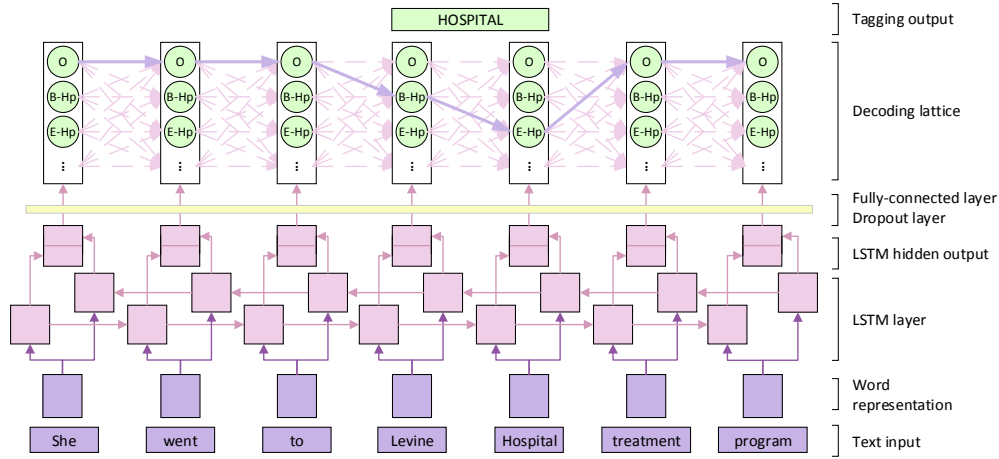

Figure 2: The architecture of LSTMs tagging model. The word-level LSTMs receive the representation of each token as input and provide the hidden layer $\overleftrightarrow{\mathbf{h}_t}$ as output. This output is used to predict the probability of tags of each word $x_t$ through a fully-connected layer, where the dropout is applied. In the decoding lattice, we find the most probable tag path from the tag lattice, along which the PHI information could be obtained.

However, this kind of combination leads to difficulty in gradient descent while training parameters, since the partial derivatives of $\mathbf{h}_t$ with respect to $\mathbf{h}_i$ continuously increase or decrease with growth in $t-i$, and finally vanish or explode [26, 27]. To address this problem, LSTMs are proposed [16]. They limit the increase in $t-i$ through a forget gate $\mathbf{f}_t$. In this study, we use a modified LSTM [28], which sets the forget gate $\mathbf{f}_t = \mathbf{1} - \mathbf{i}_t$ to reduce the parameters. The details of the gate and the cell computation of LSTM can be found in the Appendix C

The LSTM can combine the embeddings forward as well as backward. Once we obtain the forward hidden layer $\overrightarrow{\mathbf{h}_t}$ and the backward hidden layer $\overleftarrow{\mathbf{h}_t}$ and concatenate them together as $\overleftrightarrow{\mathbf{h}_t}$, we get the contextual information for this token to some extent.

### 5.2 Word representation enhancement

As the input of LSTM layer, the word embeddings of each token are pre-trained from the training data. We utilized the Word2Vec toolkit[8] and selected the skip-gram model to obtain the pre-trained, non-case-sensitive word embeddings.

Since we only used the training data, we were not able to obtain the representations of tokens that did not appear in the training set. These tokens are called out-of-vocabulary (OOV) tokens. One simple solution to this problem is to assign the embeddings of OOV tokens randomly. A better solution is to obtain the representations according to the characters composing the words [29]. Although there are hundreds of thousands of tokens in the English corpus, it consists of less than 100 characters. If we regard each token as a sequence of characters, LSTMs can be used to obtain two hidden representations of it. After concatenating the hidden representations, we get the character-level word embedding, which can represent the morphological meanings of the OOV token to some extent. Because we regard the upper and lowercase of one letter as different characters, the character-level word embeddings are case-sensitive.

---

8. https://code.google.com/archive/p/word2vec/

Moreover, although the word embeddings can be utilized directly as features, incorporating hand-crafted features is also helpful, for example, the capital and dictionary features listed in Table 2. We used two four-bit binary numbers to indicate the capital and the dictionary feature values of each word, and allocated two feature embeddings accordingly. For example, the word "Vincent" in Table 2 can be encoded as "1110" and "0010", and these two codes can be further represented as two feature embeddings.

Based on the considerations above, we concatenated the character-level word embeddings and feature embeddings to the pre-trained word embeddings to enhance the word representation. Figure 3 gives an example of the enhanced word representation. The contribution of each part of the enhanced representation is verified in Section 7.1.

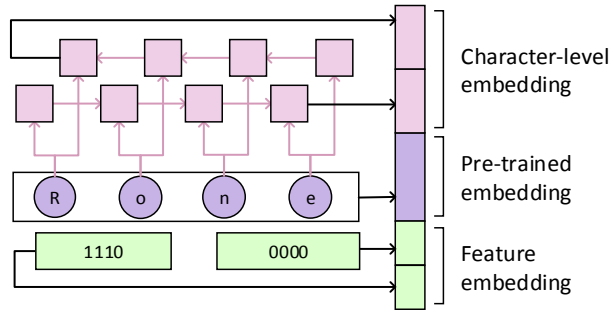

Figure 3: An example of the enhanced representation of word "Rone". It is composed of the character-level embedding, the pre-trained embedding and the feature embedding.

### 5.3 Label Decoding

Once we get hidden representation $\overleftrightarrow{\mathbf{h}_t}$ of each token $w_t$ and its context, we can predict its corresponding label immediately. That is,

$$P_t = \mathrm{softmax}(\mathbf{W}_l \overleftrightarrow{\mathbf{h}_t} + \mathbf{b}_l) \qquad (2)$$

$P_t \in \mathbb{R}^{d_l}$ and the $i$-th entry of $P_t$ is the probability that $w_t$ is labeled as the $i$-th tag. $d_l$ is the number of candidate tags. $\mathbf{W}_l \in \mathbb{R}^{d_l \times 2d_h}$ and $\mathbf{b}_l \in \mathbb{R}^{d_l}$ are the weight and bias parameters to be learned.

However, this classifier ignores the dependency of labels, which is helpful for tagging. We model this dependency in the classification layer by adding a transition matrix $\mathbf{M} \in \mathbb{R}^{d_t \times d_t}$, which is depicted as the "decoding lattice" in Figure 2. $\mathbf{M}_{ij}$ is the unnormalized transition score from the $i$-th tag to the $j$-th tag. The transition score is time invariant, which means that it is independent of the tokens. In this way, the score of a sequence $X$ and one of its predictions $y^i$ is a combination of the classification score and the transition score:

$$s(X, y^i) = \sum_{t=0}^{T-1} \mathbf{M}_{y_t^i, y_{t+1}^i} + \sum_{t=0}^{T} P_{t, y_t^i} \tag{3}$$

The probability of the $j$-th prediction $y^i$ can be obtained via a softmax classifier

$$p(y^i | X) = \frac{e^{s(X, y^i)}}{\sum_i e^{s(X, y^i)}} \tag{4}$$

During training, the log-probability of the correct label sequence is maximized:

$$\log(p(y^i | X)) = s(X, y^i) - \log(\sum_i e^{s(X, y^i)}) \tag{5}$$

Although the number of possible numbers of $y$ increases in $O(n^T)$, the calculation of Equation (5) can be completed in $O(n^2)$ time using dynamic programming. The details of the derivation can be found in the Appendix D.

In the decoding, the label sequence that obtains the maximum score is selected as the prediction:

$$y^* = \arg\max_i s(X, y^i) \tag{6}$$

## 6. Experiment

### 6.1 Corpus

The medical record set used in CEGS N-GRID 2016 Shared Task 1 in Clinical NLP contained 1,000 psychiatric evaluation records provided by Harvard Medical School. They consisted of XML documents containing raw text and PHI annotations. Table 3 shows the statistics of some main measurements of the corpus.

Table 3: Statistical overview of CEGS N-GRID 2016 Shared Task 1 corpus. The number of tokens and vocabularies were counted after pre-processing.

|  | #Notes | #Tokens | #PHI | #Unique PHI | #Vocabulary |
|---|---|---|---|---|---|
| train | 600 | 1432251 | 20845 | 14120 | 28308 |
| test | 400 | 956168 | 13519 | 9301 | 23591 |

## 6.2 Experimental setup

For CRF based system, we set the cut-off threshold of the features -f to 4, the hyper-parameter -c to 10, and used L2-norm for regularization. The hyper-parameter was determined via a subset of training data during the task. This subset is referred to from here on as the "validation set".

For LSTM-based system, the dimensions of each layer were set as follows:

- Character embedding dimension: 25

- Character-level LSTM hidden layer: 25

- Pre-trained embedding dimension: 100

- Capital embedding dimension: 6

- Dictionary embedding dimension: 6

- Word-level LSTM hidden layer: 64

Dropout [30] with a probability of 0.5 was applied to prevent overfitting. The model was trained using the stochastic gradient descent (SGD) algorithm, and the learning rate was set to 0.005. The dimension of pre-trained embedding was tuned via the validation set, while other parameters were determined through experience. The system should perform better after fine-tuning.

## 6.3 Improvement following best official run

We submitted three outputs to the task committee, the results of which are referred to from here on as the "official runs". In the experiments after the challenge, we enhanced the results further by improving the tagging models and the pre-processing module. This yielded two unofficial runs for CRFs and LSTMs.

The improvements to the models were centered around feature selection and hyper-parameter tuning. For the CRFs, we enriched the features and re-selected them in a greedy way. We also fine-tuned the hyper-parameters "-c" of the CRF++ toolkit to balance overfitting and underfitting. For LSTMs, we set the number of dimensions of the word embeddings from 50 to 100. We also added more training epochs to ensure that the parameters had been trained sufficiently.

More importantly, we added sentence detection and further improved tokenization in the pre-processing module. In the official run, we simply detected sentences according to punctuation, and many integrated sentences were hence separated, leading to a loss of contextual information. Besides, we did not tokenize the corpus in the official run as thoroughly as described in Table 1, because we thought that excessive tokenization might have split a complete PHI term into two and caused errors. In the unofficial run, we introduced the sentence detection strategy, and enriched regular expressions to ensure that as many potential entities as possible were well tokenized. The result showed that the side-effect of excessive tokenization was not severe.

The contribution of these improvements to the final performance is discussed in Section 7.1.

### 6.4 Evaluation

The system output was evaluated using precision (P), recall (R), and the $F_1$ measure, which are defined as follows:

$$P = \frac{\#(\text{true positives})}{\#(\text{true positives+false positives})} \tag{7}$$

$$R = \frac{\#(\text{true positives})}{\#(\text{true positives+false negatives})} \tag{8}$$

$$F_1 = \frac{2PR}{P+R} \tag{9}$$

According to the count method of true positives, the evaluation measures can further be classified in three independent dimensions.

There were two sets of PHI categories in the evaluation, defined by i2b2 and HIPAA. The i2b2 PHI categories were an expanded set of the HIPAA categories and contained more PHI terms, such as PROFESSION and COUNTRY.

- When the system was evaluated according to **i2b2 categories**, all subcategories under the seven main ones were evaluated.

- When the system was evaluated according to **HIPAA categories**, only the categories defined by HIPAA were evaluated.

The evaluations also differed according to whether we assessed the system at the instance level or the record level:

- **micro-F**: All of the PHI instances in the dataset were evaluated together.

- **macro-F**: Each record was evaluated and the ultimate score was obtained using the average.

If only part of the tokens of an integrated entity were identified by the system, entity-level and token-level evaluations can lead to different measures. In entity-level evaluation, there were two standards according to different matching strictness:

- **Entity level**: An entity must be identified as a whole, despite the number of tokens it contained.
  - **Strict**: The recognized entity must exactly match the first and last offsets of the gold standard entity.
  - **Relaxed**: The last offset can be off by up to 2.
- **Token level**: If the entity was separated into several parts and each was identified as the correct type, the entity was regarded as correctly identified. For example, if the PHI "2072 winter" in golden-standard with the category DATE is annotated as two DATE terms "2072" and "winter", the token-level evaluation would regard the output as correct, but the entity level would not.

Unless otherwise specified, all $F_1$ measures in the remainder of this paper are strict entity-level micro-$F_1$ measure, the primary evaluation metric in CEGS N-GRID 2016 Shared Task 1.

## 7. Results

### 7.1 Results for all categories

The best result of the three official runs mixed the outputs of the CRFs ($F_1 = 0.845$) and the LSTMs ($F_1 = 0.861$), and achieved an $F_1$ score of 0.857. The median $F_1$ score of all system outputs of the task participants was 0.822 (standard deviation = 0.183, mean = 0.779, minimum = 0.019). The best system achieved an $F_1$ of 0.914[4]. It is worth noting that the performance of the participating systems in the 2016 track was poorer than that in the 2014 track(maximum = 0.936, median = 0.845). We re-trained our participating system in 2014 track ($F_1 = 92.36\%$) [31] directly on the 2016 training set, and achieved an $F_1$ score of 82.32% on the test set. Therefore, it is reasonable to think that the de-identification on the 2016 dataset is more difficult than that on the 2014.

Our CRF-based system performed much worse on the official test set than the validation set, which drove the mixed result below that of LSTM only. This showed that the submitted CRF model, which was trained and validated by the training data, did not generalize to the test data well. With the improvements described in Section 6.3, the $F_1$ score of the CRF model increased by 3.52% and that of the LSTM model by 3.71%. Table 4 lists the evaluation measures of the best official run and the two unofficial runs based on the CRFs and the LSTMs.

Table 4: Evaluation scores for the best official run as well as two unofficial runs implemented using CRFs and LSTMs. $F$ in the table refers to micro-$F_1$ measure. The highest strict measures are shown in bold, and measures that have statistical significance compared with those of other two runs are marked with ⋆.

|  |  | Best official run | | | Unofficial CRF run | | | Unofficial LSTM run | | |
|---|---|---|---|---|---|---|---|---|---|---|
|  |  | Strict | Relaxed | Token | Strict | Relaxed | Token | Strict | Relaxed | Token |
| I2B2 | P | 0.8418 | 0.8467 | 0.8881 | **0.923** | 0.924 | 0.9463 | 0.9229 | 0.9240 | 0.9431 |
|  | R | 0.8728 | 0.8778 | 0.9087 | 0.8411 | 0.842 | 0.87 | **0.8755** | 0.8765 | 0.9015 |
|  | F | 0.857 | 0.862 | 0.8983 | 0.8802 | 0.8811 | 0.9065 | **0.8986**⋆ | 0.8996 | 0.9218 |
| HIPAA | P | 0.8767 | 0.8818 | 0.9125 | 0.9321 | 0.9333 | 0.9558 | **0.9324** | 0.9337 | 0.9509 |
|  | R | 0.9001 | 0.9054 | 0.9271 | 0.8696 | 0.8708 | 0.8696 | **0.8972** | 0.8985 | 0.9198 |
|  | F | 0.8882 | 0.8934 | 0.9197 | 0.8998 | 0.901 | 0.8998 | **0.9145**⋆ | 0.9157 | 0.9351 |

It is clear that the results of HIPAA were better than those of i2b2 because the systems performed poorly on some categories in the i2b2 set but not in the HIPAA set, such as PROFESSION, which reduced the $F_1$ score of the i2b2 categories. Further, the token-level evaluation measures were higher than the entity-level measures. It implied that the systems cannot identify the boundaries of PHI terms well in some cases, which led to extent errors.

To verify the contribution of each module of system to overall performance, we removed them separately in turn, and re-calculated the $F_1$ measure. For the CRF model, we removed one feature sub-category of five. The results are shown in Figure 4. For the LSTM model, we sequentially removed one of the three word representation parts. We also removed the dropout layer and the decoding lattice to show their influence on overall performance. When the decoding lattice was removed, the probability that the word $w_t$ was labeled using tag $t_i$ was calculated directly from Equation (2). The results are shown in Figure 5.

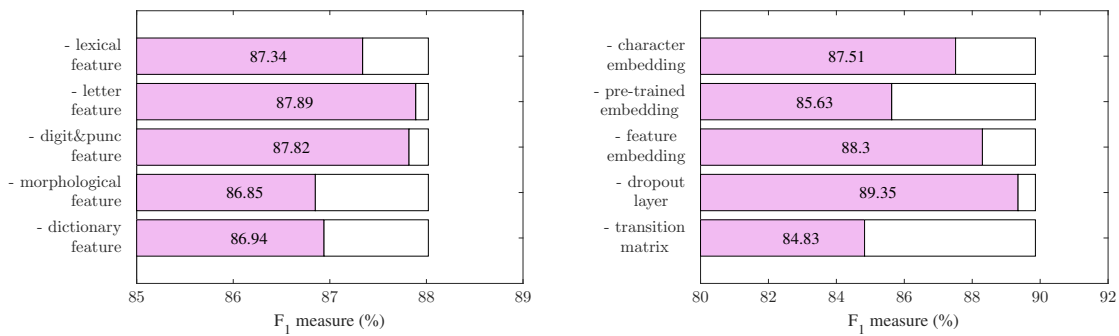

Figure 4: Overall performance curve of CRF-based system, with one feature sub-category removed from the feature set.

Figure 5: Overall performance curve of LSTM-based system, with one layer removed from the LSTM architecture.

To demonstrate the effectiveness of the improvements described in Section 6.3, we first calculated the statistical significance among the strict results of the three runs using approximate randomization[32, 33], which has been used in the last two i2b2 challenges[2, 3].

The approximate randomization method is described below. Let $\boldsymbol{o}^A = \{o_1^A, o_2^A, \cdots, o_n^A\}$ and $\boldsymbol{o}^B = \{o_1^B, o_2^B, \cdots, o_n^B\}$ denote the results of the outputs (e.g., precision, recall, and $F_1$ measure) of two systems $A$ and $B$ on $n$ test records. We calculated the difference between $\boldsymbol{o}^A$ and $\boldsymbol{o}^B$ as $d(\boldsymbol{o}^A, \boldsymbol{o}^B) = \left|\sum_i o_i^A - \sum_i o_i^B\right|$. We then randomly shuffled $\boldsymbol{o}^A$ and $\boldsymbol{o}^B$ $M$ times. For each shuffle, we randomly exchanged each $o_i^A$ and $o_i^B$ with a probability of 0.5 and obtained new $\hat{\boldsymbol{o}}^A$ and $\hat{\boldsymbol{o}}^B$. Following this, we recalculated $\hat{d}(\hat{\boldsymbol{o}}^A, \hat{\boldsymbol{o}}^B) = \left|\sum_i \hat{o}_i^A - \sum_i \hat{o}_i^B\right|$. We denoted by $m$ the number of times that $\hat{d}(\boldsymbol{o}^A, \boldsymbol{o}^B) > d(\boldsymbol{o}^A, \boldsymbol{o}^B)$, and used $\alpha = \frac{m+1}{M+1}$ to reflect the level of statistical significance. If $\alpha$ was lower than a cutoff $\bar{\alpha}$, we concluded that the results of the two systems were significant at level $\alpha$. As in the last two i2b2 tasks, we set $M = 9999$ and $\bar{\alpha} = 0.1$.

We tested the significance of the micro-averaged P, R and $F_1$ measures in Table 4. The best strict results with significant differences are marked with a star. The $\alpha$ values of both the $F_1$ measures of the unofficial LSTM run are 0.0001, which demonstrate its significant increase compared with the other two runs.

We further evaluated the individual contributions of improvements on model and pre-processing module to the increase of performance. We first modified the model and then improved the pre-processing module. The i2b2 micro-averaged $P$, $R$ and $F_1$ measures were calculated after each step. The modification of the models increased the $F_1$ measures of the two systems by 1.8% and 1.68%, and the improvement in pre-processing increased the results by another 1.69% and 2.06%. Figure 6 shows more details of the results, which suggest that pre-processing the raw text is highly beneficial before feeding it into the de-identification models.

### 7.2 Results for sub-categories

Details of the evaluation measures for each sub-category are listed in Table 5. Only the results of LSTM-based system are listed.

The ZIP category obtained the highest $F_1$ score of 100% because of its highly regular form and relatively fixed context. AGE, DATE, DOCTOR, and PHONE also achieved $F_1$ scores of more than 90% for similar reasons. Although the PATIENT sub-category belonged to the categories NAME as well as DOCTOR, its recall rate was much lower, only slightly over 70%. This was due to the high rate of missing errors, and potential causes are analyzed in Section 8.2.4.

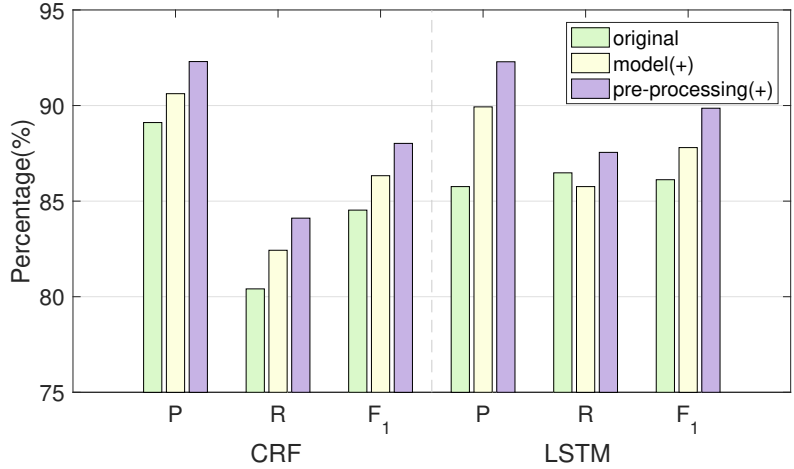

Figure 6: $P$, $R$ and $F_1$ measures change of two systems after the model and pre-processing module improvements. CRF-based system (left), LSTM-based system (right).

Table 5: The evaluation measures for each sub-category. The table lists the number of PHI instances in the training data, the gold standard data, the system output, as well as the number of instances of agreement between the last two. It also lists the P, R, and $F_1$ values of each sub-category. Only the sub-categories that appear in the training data are listed.

| Catagory | Sub-catagory | #Train | #Gold | #System | #Agree | P | R | F |
|---|---|---|---|---|---|---|---|---|
| Name | PATIENT | 1270 | 837 | 658 | 597 | 0.9073 | 0.7133 | 0.7987 |
| | DOCTOR | 2396 | 1567 | 1587 | 1491 | 0.9395 | 0.9515 | 0.9455 |
| | USERNAME | 25 | 0 | 0 | 0 | - | - | - |
| PROFESSION | | 1471 | 1010 | 828 | 688 | 0.8309 | 0.6812 | 0.7486 |
| LOCATION | HOSPITAL | 2196 | 1327 | 1211 | 1096 | 0.9050 | 0.8259 | 0.8637 |
| | ORGANIZATION | 1113 | 697 | 552 | 434 | 0.7862 | 0.6227 | 0.6950 |
| | STREET | 46 | 34 | 28 | 27 | 0.9643 | 0.7941 | 0.8710 |
| | CITY | 1394 | 820 | 892 | 748 | 0.8386 | 0.9122 | 0.8738 |
| | STATE | 662 | 481 | 459 | 419 | 0.9129 | 0.8711 | 0.8915 |
| | COUNTRY | 666 | 376 | 327 | 307 | 0.9388 | 0.8165 | 0.8734 |
| | ZIP | 23 | 17 | 17 | 17 | 1.0000 | 1.0000 | 1.0000 |
| | LOCATION-OTHER | 25 | 19 | 13 | 4 | 0.3077 | 0.2105 | 0.2500 |
| AGE | | 3637 | 2354 | 2317 | 2234 | 0.9642 | 0.9490 | 0.9565 |
| DATE | | 5723 | 3821 | 3790 | 3646 | 0.9620 | 0.9542 | 0.9581 |
| CONTACT | PHONE | 143 | 113 | 112 | 106 | 0.9464 | 0.9381 | 0.9422 |
| | FAX | 4 | 5 | 3 | 3 | 1.0000 | 0.6000 | 0.7500 |
| | EMAIL | 2 | 5 | 1 | 0 | 0 | 0 | 0 |
| ID | URL | 5 | 3 | 1 | 0 | 0 | 0 | 0 |
| | MEDICALRECORD | 4 | 2 | 1 | 0 | 0 | 0 | 0 |
| | HEALTHPLAN | 4 | 2 | 0 | 0 | 0 | 0 | 0 |
| | LICENSE | 38 | 21 | 28 | 19 | 0.6786 | 0.9048 | 0.7755 |
| | IDNUM | 2 | 8 | 0 | 0 | 0 | 0 | 0 |

EMAIL and ID were also regular, but instances of these categories were rare compared with the above categories. A data-driven system was unlikely to learn patterns from so few instances, and thus yielded poor performance. For these categories, regular expressions may be helpful.

Three classes of sub-categories were shown in the LOCATION category. The first contained CITY, STATE, and COUNTRY, which had relatively fixed dictionaries, and obtained higher $F_1$ scores. The second contained HOSPITAL and STREET. These sub-categories had some signal words, such as "hospital," "clinic," "road," and "avenue," which were also helpful for recognition. The third class, which contained ORGANIZATION and LOCATION-OTHER, had neither complete dictionaries nor signal words, and thus was more difficult to recognize.

PROFESSION was one of the most difficult categories to identify. One reason for this was the lack of dictionaries and signal words, and another was that the contexts for the terms in this category were complicated. Moreover, the terms in PROFESSION could be long, such as "Telecommunications Installation and Repair Worker" and "Inspector in Public and Environmental Health and Occupational Health and Safety." Identification of such terms needed to rely on longer contextual information, which is still a hard problem in NLP.

## 8. Discussion

### 8.1 Comparison of CRFs and LSTMs

LSTMs outperform CRFs in all $F_1$ measures. LSTM-based systems have similar precision but much higher recall than CRF-based systems. This is because the hand-crafted features used in CRFs are selected carefully for token classification, but cannot cover all scenarios. On the contrary, the automatic features utilized by LSTMs are derived directly from data, and can depict intrinsic features hidden in the data. Thus, the features are more general and the recall is much higher than that of CRFs.

From a model-building perspective, LSTMs have an additional hidden layer compared with CRFs, which have only two layers. If we regard LSTMs as deep neural networks, CRF is a kind of shallow network. It has been claimed that deep neural networks have more powerful fitting capabilities compared with shallow log-linear models. Moreover, LSTM can model long-term dependency, which can capture contextual information for a longer period.

An advantage of CRFs is that they optimize entire sequences of tags rather than tags of each token. To use this in the LSTMs, we introduce the transition matrix while decoding the labels.

## 8.2 Error analysis

### 8.2.1 Error categories

Errors under the entity-level strict evaluation were divided into four categories according to [34]:

- Type error: The entity was identified by a correct start and end location but the wrong type.

- Extent error: The location span of the entity overlapped with that of a gold-standard entity, but did not match it exactly. There were three scenarios of overlapping:

  - Short: The location span of the entity fell within that of a gold-standard entity.
  - Long: The location span of the entity covered that of a gold-standard entity.
  - Short&Long: The location span of the entity neither fell into nor covered that of a gold-standard entity

- Spurious error: The location span of the entity had no overlap with any correct entity.

- Missing error: The location span of the entity in the gold standard had no overlap with that of any entity in the system.

Figure 7 shows the distribution of the four error categories. The percentages of type error, spurious error, and extent error were calculated based on the system output, whereas the percentages of missing error were calculated based on gold-standard data. If a PHI term produced by the system was incorrect, the error would certainly be grouped into one of the above four categories, except missing error. Consider the first column of Figure 7 as an example. It shows that approximately 91% of the entities in PATIENT were identified by system according to the gold standard, 2% of which from the category DOCTOR, 4% were not PHI terms, and 1% were shorter than their corresponding standard answers. The sum of these percentages is not exactly 100% because the values in each block were rounded due to limitations of space. The original data for this figure can be found in Table A.1. From the first row of missing errors in the sub-figure, we see that approximately 23% of the terms in PATIENT in the gold-standard were not identified by the system.

### 8.2.2 Type errors

Type errors are shown as a confusion matrix in Figure 7. It can be seen easily that entities tended to be identified incorrectly among the sub-categories belonging to a main category. For example, 19 entities of PATIENT were incorrectly identified as entities of DOCTOR, and 14 entities of DOCTOR were incorrectly identified as those of PATIENT. The same scenario also obtained among sub-categories in LOCATION and CONTACT. The difficulty in finer-grained classification lay in the similarity of the morphological features and the context of these entities.

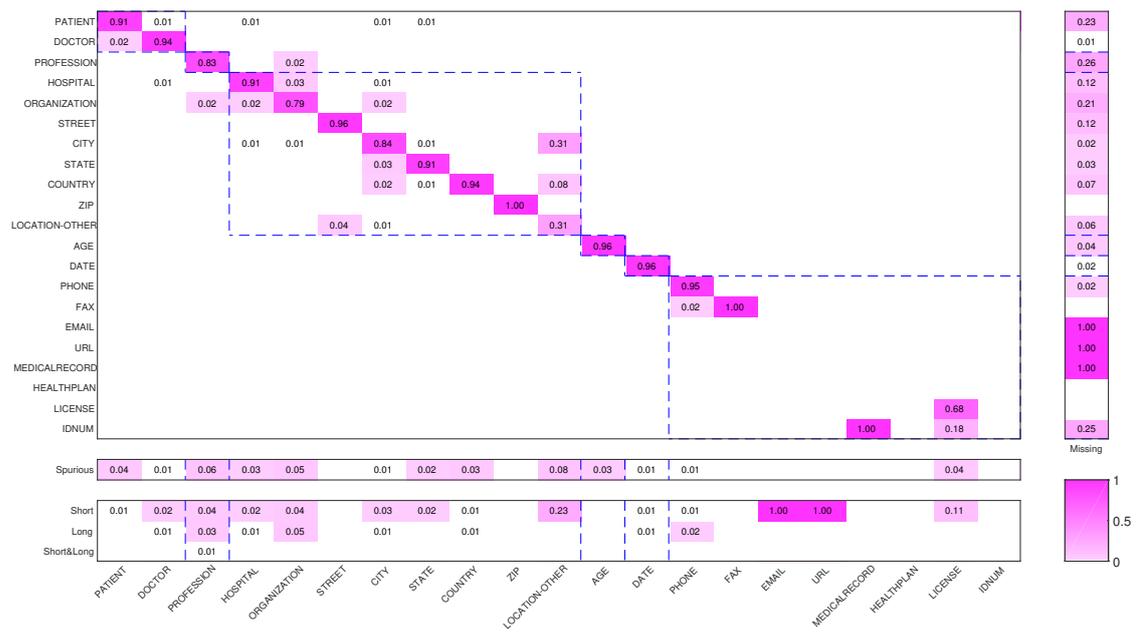

Figure 7: The visualization of error distribution. The main sub-figure is a confusion matrix used to depict type error. The other three sub-figures show the distribution of missing, spurious, and extent errors. The meanings of each row and column are listed in the headings. Different PHI categories are separated by the dashed blue line.

There were also several confusions between the categories PROFESSION and LOCATION. This was because they sometimes occurred together and shared similar contexts. For example, the sentence "a retired Landscape architecht from Albemarle Corporation" contains a PROFESSION term (Landscape architecht) and an ORGANIZATION term (Albemarle Corporation).

### 8.2.3 Spurious errors

Spurious errors occurred when ordinary tokens had similar lexical or contextual features with real PHI entities. For example, a token with the first letter capitalized was likely to be identified as a PHI term, and one consisting of two digits tended to be recognized as DATE or AGE. In other cases, the word itself had more than one semantic meaning. For example, in the sentence "Her last depressive episode was last winter," the token "winter" was annotated as a DATE entity. However, in the phrase "winter boots," the token was just an adjective rather than a PHI term. There were also some cases where the tokens themselves were confusing. For example, the system identified "big company" and "JEwish day schools" as ORGANIZATION, though they were not exactly PHI terms.

### 8.2.4 Missing errors

Missing errors occurred more often in sub-categories with few instances in the training set. For EMAIL, URL, and MEDICALRECORD, there were fewer than 10 instances each, and it was hence natural that the error rate in the test set was high. In addition to CONTACT, PROFESSION and ORGANIZATION had the highest missing rates. As discussed above, this was due to a lack of complete domain dictionaries and signal words.

It is interesting that although both PATIENT and DOCTOR were sub-categories of NAME, the system yielded entirely different missing rate for these two sub-categories. The missing rate of entities in PATIENT was approximately 23%, much higher than the 1% of those in DOCTOR. In addition to the more complicated context of PATIENT, we found that the document frequency of PATIENT was much lower than that of DOCTOR. This meant that the system more easily remembered token of DOCTOR. The document frequency distributions of the main sub-categories are shown in Figure 8.

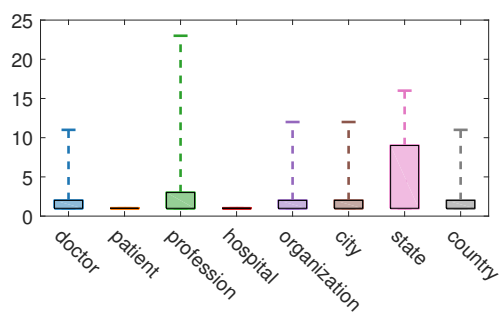

Figure 8: The boxplot of document frequencies of the main sub-categories.

8.2.5 Extent errors

The occurrence of extent errors showed that the model could not satisfactorily detect the boundaries of several entities. We checked cases for all three types of extent errors and described them below.

There were two main causes of short errors. The first was tokenization. For example, "Zenith Uni." was an ORGANIZATION entity. However, since the period was separated from "Uni" during tokenization, the system can only identify "Zenith Uni" as a PHI term. Another type of short error was that the system tags one entity as two. "State University of Wyoming" was an ORGANIZATION entity. However, the system tagged "State University" as an ORGANIZATION and "Wyoming" as a STATE. This showed that the system could not handle the "of" phrase structure well. It sometimes tagged two sub-structures of an integrated term with "of" as two independent entities. Similar examples were "winter of 2091" and "Cancer Center of America."

Similarly, these two problems can also cause long errors. For example, the system tagged "Educare-Fargo\," rather than "Educare-Fargo" as a "hospital" entity because of the failure of tokenizing "\" out from "Fargo." The system sometimes also tagged two or more entities as one. For example, the entity "computer science health informatic" identified by the system were in fact two entities in the gold standard. There were some long errors as well that were confusing. For example, the system tagged "**Woodland Park** High School" as an "organization," but the gold standard tagged only "Woodland Park" as a "city." Similar examples were "**landscaping** employer" and "**HMC** Home Services." Perhaps some extra information, like tf-idf, can help the system filter more general tokens, such as "high school" and "employer."

Short&Long errors occurred less frequently compared with the above two extent errors, and mainly in PROFESSION when preceded by an ORGANIZATION entity. For example, in the sentence "33yo married palauan female Bob Evans buildings construction worker," the gold standard tagged "Bob Evans buildings" and "construction worker" as ORGANIZATION and PROFESSION, respectively. However, the system tagged "buildings construction worker" as a PROFESSION entity.

### 8.3 Limitations

Text in the clinical domain has its own characteristics. The direct transformation of NER from the open to the clinical domain is not the best way. However, limited by the time available for the task, specific processing against clinical narratives was not introduced to the systems. The corpus was not exploited fully, and only a small part of the dictionaries was utilized. The sentence detector and the POS/chunk tagger were open-domain toolkits. We also did not add a post-processing module. We believe these domain-specific toolkits, rules, and resources can further improve the performance of the system.

At the same time, open-domain prior knowledge is also helpful for some categories, such as PROFESSION and LOCATION, which are also common named entities in open-domain corpora. The pre-trained word embedding based on these large-scale corpora should be helpful for identification.

## 9. Conclusion

This paper proposed two automatic de-identification systems based on CRFs and LSTMs. The LSTM-based system attained a micro-$F_1$ score of 89.86% in i2b2 strict evaluation, which was higher than that of the CRF-based system. LSTMs can identify PHI terms without depending on hand-crafted features and obtain higher recall rates than CRFs. In addition to the model, the pre-processing module can significantly affect the performance of the system. Accurate sentence detection and tokenization is a premise and foundation of subsequent PHI term recognition.

Furthermore, as Section 8.3 pointed out, we will attempt to incorporate prior knowledge and domain-specific resources to help increase the results for categories that yielded poor performance.

### Acknowledgment

The CEGS N-GRID 2016 Shared Task 1 in Clinical Natural Language Processing was supported by NIH P50 MH106933, NIH 4R13LM011411. This work is also supported by the Natural Science Foundation of China (No. 71531007). The authors would like to thank the organizing committee for this task and the annotators of the dataset. We also thank the anonymous reviewers for their comments, which provided us with significant guidance.

# Appendix A.

Table A.1 is a quantitative version of Figure 7.

Table A.1: Error distribution of system output.

|  | system output | | | | | | | | | | | | | | | | | | | | |
|---|---|---|---|---|---|---|---|---|---|---|---|---|---|---|---|---|---|---|---|---|---|
|  | Pt | Dct | Pf | Hpt | Og | Strt | Ct | Stat | Ct | Zip | L-O | Age | Dt | Phn | Fax | Em | Url | Mrd | Hp | Lcs | ID | Missing | Total |
| PATIENT | 597 | 19 | 1 | 8 | 2 | | 5 | 6 | 1 | | | | | | | | | | | | | 187 | 826 |
| DOCTOR | 14 | 1491 | | 4 | 2 | | 4 | | | | | | 1 | | | | | | | | | 22 | 1538 |
| PROFESSION | 1 | | 688 | | 9 | | | | | | | | | | | | | | | | | 242 | 940 |
| HOSPITAL | 3 | 11 | | 1096 | 19 | | 13 | 2 | | | | | | | | | | | | | | 151 | 1295 |
| ORGANIZATION | 2 | | 18 | 26 | 434 | | 14 | 1 | 1 | | | 1 | | | | | | | | | | 136 | 633 |
| STREET | | 1 | | | | 27 | 1 | | | | | | | | | | | | | | | 4 | 33 |
| CITY | 1 | 1 | 2 | 11 | 5 | | 748 | 4 | 1 | | 4 | | | | | | | | | | | 18 | 795 |
| STATE | | 2 | | | | | 30 | 419 | 1 | | | 1 | 3 | | | | | | | | | 14 | 470 |
| COUNTRY | 3 | | | 1 | 2 | | 22 | 5 | 307 | | 1 | | | | | | | | | | | 26 | 367 |
| ZIP | | | | | | | | | | 17 | | | | | | | | | | | | 0 | 17 |
| LOCATION-OTHER | | | | 1 | 1 | 1 | 10 | | | | 4 | | | | | | | | | | | 1 | 18 |
| AGE | | | | | | | | 2 | | | | 2234 | 5 | | | | | | | | | 97 | 2338 |
| DATE | 2 | 1 | 1 | | 2 | | 1 | 1 | | | | 11 | 3646 | | | | | | | | | 56 | 3721 |
| PHONE | | | | 1 | | | | | | | | | | 106 | | | | | | | | 2 | 109 |
| FAX | | | | | | | | | | | | | | 2 | 3 | | | | | | | 0 | 5 |
| EMAIL | | | | | | | | | | | | | | | | 0 | | | | | | 4 | 4 |
| URL | | | | | | | | | | | | | | | | | 0 | | | | | 2 | 2 |
| MEDICALRECORD | | | | | | | | | | | | | | | | | | 0 | | | | 2 | 2 |
| HEALTHPLAN | | | | | | | | | | | | | | | | | | | 0 | | | 0 | 0 |
| LICENSE | | | | | | | | | | | | | | | | | | | | 19 | | 0 | 19 |
| IDNUM | | | | | | | | | | | | | | | | | | | 1 | 5 | 0 | 2 | 8 |
| total | 623 | 1526 | 710 | 1148 | 476 | 28 | 848 | 440 | 311 | 17 | 9 | 2247 | 3655 | 108 | 3 | 0 | 0 | 1 | 0 | 24 | 0 | 966 | 13140 |
| Spurious | 26 | 20 | 52 | 33 | 29 | | 9 | 8 | 11 | | 1 | 59 | 45 | | | | | | | 1 | | | 294 |
| short | 9 | 27 | 35 | 21 | 21 | | 26 | 10 | 3 | | 3 | 6 | 52 | 1 | | 1 | 1 | | | 3 | | 1 | 220 |
| long | | 14 | 26 | 9 | 26 | | 9 | 1 | 2 | | | 5 | 37 | 2 | | | | | | | | | 131 |
| sl | | | 5 | | | | | | | | | | 1 | | | | | | | | | 5 | 11 |
| total | 9 | 41 | 66 | 30 | 47 | 0 | 35 | 11 | 5 | 0 | 3 | 11 | 90 | 3 | 0 | 1 | 1 | 0 | 0 | 3 | 0 | 6 | 362 |
| system | 658 | 1587 | 828 | 1211 | 552 | 28 | 892 | 459 | 327 | 17 | 13 | 2317 | 3790 | 112 | 3 | 1 | 1 | 1 | 0 | 28 | 0 | 973 | |

# Appendix B.

When the CRF is applied to NER, $X = \mathbf{x}_0, \mathbf{x}_1, \cdots, \mathbf{x}_t, \cdots, \mathbf{x}_T$ denotes the features of an input sentence of length $T$, where $\mathbf{x}_t$ is the features of the word $x_t$ at position $t$ and its context. And $Y = y_1, y_2, \cdots, y_t, \cdots, y_T$ denotes the corresponding output labels of each word $x_t$.

The feature set used in linear-chain CRF can be written as $\mathcal{F} = \{f_k(y_t, y_{t-1}, \mathbf{x}_t) | \forall k\}$, where $f_k(y_t, y_{t-1}, \mathbf{x}_t) = \boldsymbol{q}_k(\mathbf{x}_t)\mathbf{I}(y_t = y)\mathbf{I}(y_{t-1} = y')$. $\boldsymbol{q}_k(\mathbf{x}_t)$ is the $k$-th observed feature of $x_t$ and its context, and $\mathbf{I}(\cdot)$ is the indicator function. It can further simplified as two kinds of features: $f_k(y_t, \mathbf{x}_t) = \boldsymbol{q}_k(\mathbf{x}_t)\mathbf{I}(y_t = y)$ and $f(y_t, y_{t-1}) = \mathbf{I}(y_t = y)\mathbf{I}(y_{t-1} = y')$.

Then, the distribution of $Y$ given $X$ can be written as

$$p(Y|X) = \frac{1}{Z(X)} \prod_{t=1}^{T} \Psi_t(y_t, y_{t-1}, \mathbf{x}_t) \tag{10}$$

where

$$\Psi_t(y_t, y_{t-1}, \mathbf{x}_t) = \exp\{\sum_{k=1}^{K} \theta_k f_k(y_t, y_{t-1}, \mathbf{x}_t)\} \tag{11}$$

is the log-linear combination of the feature space. $\theta_k$ is the corresponding weight parameter of $k$-th feature $f_k(\cdot)$, which can be learned by regularized maximum-likelihood estimation (MLE). $K = |\mathcal{F}|$ is the size of feature set.

## Appendix C.

LSTMs introduce cell state $\mathbf{c}_t$ to cover all information over time. At every time step, $\mathbf{c}_t$ is updated with $\mathbf{c}_{t-1}$ and $\mathbf{z}_t$, which is exactly the same as in Equation (1). There are two gates, $\mathbf{i}_t$ and $\mathbf{o}_t$, which are calculated by the weight combination of current embedding $\mathbf{x}_t$, last $\mathbf{h}_{t-1}$ and last $\mathbf{c}_{t-1}$, and output through a sigmoid function. This guarantees that each element of $\mathbf{i}_t$, $\mathbf{1} - \mathbf{i}_t$, and $\mathbf{o}_t$ are numbers in $[0, 1]$. After pointwise multiplication operation with another vector with the same shape, they determine the contribution of this vector to the result. In LSTMs, $\mathbf{i}_t$ modulates the combination of $\mathbf{c}_{t-1}$ and $\mathbf{z}_t$, whereas $\mathbf{o}_t$ modulates the contribution of $\mathbf{c}_t$ to $\mathbf{h}_t$:

$$\mathbf{z}_t = \tanh(\mathbf{W}_z \mathbf{x}_t + \mathbf{U}_z \mathbf{h}_{t-1} + \mathbf{b}_z) \tag{12}$$

$$\mathbf{i}_t = \sigma(\mathbf{W}_i \mathbf{x}_t + \mathbf{U}_i \mathbf{h}_{t-1} + \mathbf{V}_i \mathbf{c}_{t-1} + \mathbf{b}_i) \tag{13}$$

$$\mathbf{c}_t = (\mathbf{1} - \mathbf{i}_t)\mathbf{c}_{t-1} + \mathbf{i}_t \mathbf{z}_t \tag{14}$$

$$\mathbf{o}_t = \sigma(\mathbf{W}_o \mathbf{x}_t + \mathbf{U}_o \mathbf{h}_{t-1} + \mathbf{V}_o \mathbf{c}_{t-1} + \mathbf{b}_o) \tag{15}$$

$$\mathbf{h}_t = \mathbf{o}_t \odot \tanh(\mathbf{c}_t) \tag{16}$$

$\mathbf{W} \in \mathbb{R}^{d_h \times d_e}$, $\mathbf{U} \in \mathbb{R}^{d_h \times d_h}$, $\mathbf{V} \in \mathbb{R}^{d_h \times d_h}$, and $\mathbf{b} \in \mathbb{R}^{d_h}$ are weight and bias parameters to be learned. These parameters are time invariant, and reduce the size of the hypothesis space.

## Appendix D.

We present the formula used to calculate $\log(\sum_i e^{s(X, y^i)})$ in Equation (5). This term is a form of log-sum-exp, and we rewrite it as $\underset{\forall y_{1:T}}{\mathrm{LSE}}\, s(X_{1:T}, y_{1:T})$. Like the forward algorithm in CRF, a middle variable $\delta_t(k)$ is introduced:

$$\begin{aligned}
\delta_t(k) &\triangleq \operatorname*{LSE}_{\forall y_{1:t} \cap y_t = k} s(X_{1:t}, y_{1:t}) \\
&= \operatorname*{LSE}_{\forall i, y_{1:t} \cap y_t = k} (s(X_{1:t}, y_{1:t-1})|_{y_{t-1}=i} + \mathbf{M}_{ik} + P_{t,k}) \\
&= \operatorname*{LSE}_{\forall i, y_{1:t} \cap y_t = k} (s(X_{1:t}, y_{1:t-1})|_{y_{t-1}=i} + \mathbf{M}_{ik}) + P_{t,k} \\
&= \operatorname*{LSE}_{\forall i, y_t = k} (\operatorname*{LSE}_{\forall y_{1:t-1} \cap y_{t-1} = i} [s(X_{1:t-1}, y_{1:t-1}) + \mathbf{M}_{ik}]) + P_{t,k} \\
&= \operatorname*{LSE}_{\forall i, y_t = k} (\operatorname*{LSE}_{\forall y_{1:t-1} \cap y_{t-1} = i} s(X_{1:t-1}, y_{1:t-1}) + \mathbf{M}_{ik}) + P_{t,k} \\
&= \operatorname*{LSE}_{\forall i, y_t = k} (\delta_{t-1}(i) + \mathbf{M}_{ik}) + P_{t,k}
\end{aligned}$$

Finally, we obtain $\log(\sum_i e^{s(X, y^i)}) = \operatorname*{LSE}_{\forall y_{1:T}} s(X_{1:T}, y_{1:T}) = \sum_k \delta_T(k)$.

# References

[1] T D Gunter and N P Terry. The emergence of national electronic health record architectures in the United States and Australia: models, costs, and questions. *Journal of Medical Internet Research*, 7(1):e3, 2005.

[2] Özlem Uzuner, Yuan Luo, and Peter Szolovits. Evaluating the state-of-the-art in automatic de-identification. *Journal of the American Medical Informatics Association*, 14(5):550–563, 2007.

[3] Amber Stubbs, Christopher Kotfila, and Özlem Uzuner. Automated systems for the de-identification of longitudinal clinical narratives: Overview of 2014 i2b2/UTHealth shared task Track 1. *Journal of Biomedical Informatics*, 58:S11—-S19, 2015.

[4] Amber Stubbs, Michele Filannino, and Özlem Uzuner. De-identification of psychiatric intake records: Overview of 2016 CEGS N-GRID shared tasks Track 1. *Journal of Biomedical Informatics*, 2017.

[5] Stéphane Meystre, Shuying Shen, Deborah Hofmann, and Adi Gundlapalli. Can Physicians Recognize Their Own Patients in De-identified Notes? In *Studies in Health Technology and Informatics*, volume 205, pages 778–782, 2014.

[6] Cyril Grouin, Rue John von Neuman, Nicolas Griffon, and Aurélie Névéol. Is it possible to recover personal health information from an automatically de-identified corpus of french ehrs? In *Sixth International Workshop On Health Text Mining And Information Analysis (Louhi)*, page 31, 2015.

[7] Ralph Grishman and Beth Sundheim. Message Understanding Conference-6: A Brief History. In *COLING*, volume 96, pages 466–471, 1996.

[8] GuoDong Zhou and Jian Su. Named entity recognition using an HMM-based chunk tagger. In *Proceedings of the 40th Annual Meeting on Association for Computational Linguistics*, pages 473–480. Association for Computational Linguistics, 2002.

[9] Andrew McCallum and Wei Li. Early results for named entity recognition with conditional random fields, feature induction and web-enhanced lexicons. In *Proceedings of the seventh conference on Natural language learning at HLT-NAACL 2003-Volume 4*, pages 188–191. Association for Computational Linguistics, 2003.

[10] Hideki Isozaki and Hideto Kazawa. Efficient support vector classifiers for named entity recognition. In *Proceedings of the 19th international conference on Computational linguistics-Volume 1*, pages 1–7. Association for Computational Linguistics, 2002.

[11] Thomas G Dietterich. Machine learning for sequential data: A review. In *Joint IAPR International Workshops on Statistical Techniques in Pattern Recognition (SPR) and Structural and Syntactic Pattern Recognition (SSPR)*, pages 15–30. Springer, 2002.

[12] Hui Yang and Jonathan M Garibaldi. Automatic detection of protected health information from clinic narratives. *Journal of Biomedical Informatics*, 58:S30—-S38, 2015.

[13] Yoshua Bengio, Aaron Courville, and Pascal Vincent. Representation learning: A review and new perspectives. *IEEE transactions on pattern analysis and machine intelligence*, 35(8):1798–1828, 2013.

[14] Yann LeCun, Yoshua Bengio, and Geoffrey Hinton. Deep learning. *Nature*, 521(7553):436–444, 2015.

[15] Ronan Collobert, Jason Weston, Léon Bottou, Michael Karlen, Koray Kavukcuoglu, and Pavel Kuksa. Natural language processing (almost) from scratch. *Journal of Machine Learning Research*, 12(Aug):2493–2537, 2011.

[16] Sepp Hochreiter and Jürgen Schmidhuber. Long short-term memory. *Neural computation*, 9(8):1735–1780, 1997.

[17] Zhiheng Huang, Wei Xu, and Kai Yu. Bidirectional LSTM-CRF models for sequence tagging. *arXiv preprint arXiv:1508.01991*, 2015.

[18] Jason P C Chiu and Eric Nichols. Named entity recognition with bidirectional LSTM-CNNs. *Transactions of the Association for Computational Linguistics*, 4:357–370, 2016.

[19] Xuezhe Ma and Eduard Hovy. End-to-end Sequence Labeling via Bi-directional LSTM-CNNs-CRF. *arXiv preprint arXiv:1603.01354*, 2016.

[20] Guillaume Lample, Miguel Ballesteros, Sandeep Subramanian, Kazuya Kawakami, and Chris Dyer. Neural architectures for named entity recognition. *arXiv preprint arXiv:1603.01360*, 2016.

[21] Franck Dernoncourt, Ji Young Lee, Ozlem Uzuner, and Peter Szolovits. De-identification of Patient Notes with Recurrent Neural Networks. *arXiv preprint arXiv:1606.03475*, 2016.

[22] Laura Chiticariu, Yunyao Li, and Frederick R Reiss. Rule-Based Information Extraction is Dead! Long Live Rule-Based Information Extraction Systems! In *EMNLP*, number October, pages 827–832, 2013.

[23] Jonathon Read, Rebecca Dridan, Stephan Oepen, and Lars Jørgen Solberg. Sentence Boundary Detection : A Long Solved Problem ? *Coling 2012*, (December 2012):985–994, 2012.

[24] Tomas Mikolov, Kai Chen, Greg Corrado, and Jeffrey Dean. Efficient estimation of word representations in vector space. *arXiv preprint arXiv:1301.3781*, 2013.

[25] John Lafferty, Andrew McCallum, and Fernando Pereira. Conditional random fields: Probabilistic models for segmenting and labeling sequence data. In *Proceedings of the eighteenth international conference on machine learning, ICML*, volume 1, pages 282–289, 2001.

[26] Razvan Pascanu, Tomas Mikolov, and Yoshua Bengio. On the difficulty of training recurrent neural networks. *ICML (3)*, 28:1310–1318, 2013.

[27] Yoshua Bengio, Patrice Simard, and Paolo Frasconi. Learning long-term dependencies with gradient descent is difficult. *IEEE transactions on neural networks*, 5(2):157–166, 1994.

[28] Klaus Greff, Rupesh Kumar Srivastava, Jan Koutn\'\ik, Bas R Steunebrink, and Jürgen Schmidhuber. LSTM: A search space odyssey. *arXiv preprint arXiv:1503.04069*, 2015.

[29] Xiang Zhang, Junbo Zhao, and Yann LeCun. Character-level convolutional networks for text classification. In *Advances in neural information processing systems*, pages 649–657, 2015.

[30] Nitish Srivastava, Geoffrey E Hinton, Alex Krizhevsky, Ilya Sutskever, and Ruslan Salakhutdinov. Dropout: a simple way to prevent neural networks from overfitting. *Journal of Machine Learning Research*, 15(1):1929–1958, 2014.

[31] Bin He, Yi Guan, Jianyi Cheng, Keting Cen, and Wenlan Hua. CRFs based de-identification of medical records. *Journal of biomedical informatics*, 58:S39—-S46, 2015.

[32] Nancy Chinchor. The Statistical Significance of the MUC-4 Results. In *Proceedings of the 4th Conference on Message Understanding*, pages 30—-50, 1992.

[33] E W Noreen. *Computer-intensive methods for testing hypotheses: an introduction*. 1989.

[34] B Wellner, M Huyck, S Mardis, J Aberdeen, A Morgan, L Peshkin, A Yeh, J Hitzeman, and L Hirschman. Rapidly retargetable approaches to de-identification in medical records. *Journal of the American Medical Informatics Association*, 14(5):564–573, 2006.